\documentclass[12pt, reqno]{zapiski}
\originfo{529}{}{}{2026}
\usepackage[T2A]{fontenc}
\usepackage[utf8]{inputenc}
\usepackage{graphicx}
\usepackage{amsmath}
\usepackage{hyperref}
\hypersetup{
  colorlinks=true,
  linkcolor=blue,
  urlcolor=blue,
  citecolor=blue,
  pdfborder={0 0 0},
  hypertexnames=false
}
\usepackage{xcolor}
\usepackage{array}
\usepackage{longtable}
\usepackage{calc}
\usepackage{booktabs}
\usepackage{float}
\usepackage[section]{placeins}

\providecommand{\tightlist}{%
  \setlength{\itemsep}{0pt}\setlength{\parskip}{0pt}}

\newcommand{\fittable}[1]{%
  \setbox0=\hbox{#1}%
  \ifdim\wd0>\textwidth
    \resizebox{\textwidth}{!}{#1}%
  \else
    \centering#1%
  \fi
}

\newlength{\tablecaptionskip}
\makeatletter
\long\def\@makecaption#1#2{%
  \setbox\@tempboxa\vbox{\color@setgroup
    \advance\hsize-2\captionindent\noindent
    \@captionfont\@captionheadfont#1\@xp\@ifnotempty\@xp
        {\@cdr#2\@nil}{.\@captionfont\upshape\enspace#2}%
    \unskip\kern-2\captionindent\par
    \global\setbox\@ne\lastbox\color@endgroup}%
  \ifhbox\@ne
    \setbox\@ne\hbox{\unhbox\@ne\unskip\unskip\unpenalty\unkern}%
  \fi
  \ifdim\wd\@tempboxa=\z@
    \setbox\@ne\hbox to\columnwidth{\hss\kern-2\captionindent\box\@ne\hss}%
  \else
    \setbox\@ne\vbox{\unvbox\@tempboxa\parskip\z@skip
        \noindent\unhbox\@ne\advance\hsize-2\captionindent\par}%
  \fi
  \ifnum\@tempcnta<64
    \addvspace\abovecaptionskip
    \hbox to\hsize{\kern\captionindent\box\@ne\hss}%
  \else
    \vskip\tablecaptionskip
    \hbox to\hsize{\kern\captionindent\box\@ne\hss}%
    \nobreak
    \vskip\belowcaptionskip
  \fi
\relax}
\makeatother

\english
\def\publnamef{\hbox{}}
\def\publnames{\hbox{}}
\def\russianvolinfo{\hbox{}}
\renewcommand{\datename}{Received}
\makeatletter
\def\@serieslogo{}
\def\@setdate{}
\makeatother

\makeatletter
\def\@setabstracta{%
  \ifvoid\abstractbox
  \else
    \skip@18\p@ \advance\skip@-\lastskip
    \advance\skip@-\baselineskip \vskip\skip@
    \unvbox\abstractbox
    \prevdepth\z@
  \fi
}
\makeatother

\title[VLM Anchoring Bias]{Don't Look at the Numbers: Visual Anchoring Bias and Layer-wise Representation in VLMs}
\author[M.\,Shalankin]{M.\,Shalankin}
\date{April 2026}

\begin{document}
\sloppy
\begin{abstract}
Embedded numeric anchors on images systematically bias Vision-Language Model quality judgments across six VLMs from five architectural families (ANOVA $\eta^2$ = 0.18--0.77, all p < 0.001). Anchor effects are 2.5$\times$ larger than severe image quality degradation, confirming bias is not reducible to visual changes. Layer-wise probing reveals consistent dissociation: layers where anchor classification saturates (L12--L34) are suboptimal for quality prediction, with optimal layers deeper ($R^2$ = 0.69--0.91). Fusion analysis identifies architecture-dependent integration---instant fusion at L1--L2 in two models versus partial or no fusion in three others. These results establish a causal account of visual anchoring bias, linking behavioral susceptibility to representation dynamics.
\end{abstract}


\maketitle

\section{Introduction}\label{introduction}

\subsection{Motivation}\label{motivation}

Vision-Language Models (VLMs) are increasingly used for image quality assessment and evaluation tasks, yet their robustness to embedded text manipulation remains poorly understood. Prior work has examined adversarial attacks and typographic manipulation in classification tasks, but a critical gap exists: how do text overlays affect continuous judgment tasks such as visual quality assessment? If a model's quality rating can be manipulated simply by superimposing a numeric anchor on an image, this raises concerns about the reliability of VLM-based evaluation systems.

Recent advances in mechanistic interpretability have begun to localize where VLMs process text~\cite{steinberg2026,shi2025}, but these findings address where text is ``read''---not how text influences subsequent judgments. The relationship between OCR capability and cognitive bias induction remains unexplored.

\subsection{Research Questions}\label{research-questions}

This work addresses five interconnected questions:

\textbf{RQ1:} How strongly do systematic text anchors (numeric ratings overlaid on images) influence VLM quality judgments?

\textbf{RQ2:} Is there a correlation between anchor value and model rating, and what is its magnitude?

\textbf{RQ3:} In which layers of the transformer architecture does text reading capability emerge?

\textbf{RQ4:} Do layers optimal for text classification coincide with layers optimal for visual quality representation?

\textbf{RQ5:} Does chain-of-thought reasoning provide robustness against anchor influence?

\subsection{Contributions}\label{contributions}

Our work makes five primary contributions:

\textbf{1. Cross-architecture quantification of anchor susceptibility with causal evidence.} We demonstrate significant anchor susceptibility across six VLMs from five architectural families (ANOVA $\eta^2$ = 0.18--0.77; all Wilcoxon p < 0.001), with a 4.3$\times$ effect magnitude range between the most susceptible (Qwen3-VL-8B, $\eta^2$ = 0.77) and most resistant (Gemma-4-E4B, $\eta^2$ = 0.18) models. Per-image delta analysis (Cohen's d up to 3.35) establishes that these effects are causal, not merely correlational.

\textbf{2. Identification of saturation and optimal layers across five architectures.} Layers where text classification saturates (L12--L34 across models) are suboptimal for visual quality prediction. Optimal quality prediction occurs at deeper layers with $R^2$ = 0.69--0.91, revealing a dissociation between ``reading text'' and ``representing visual quality'' that is consistent across all probed architectures.

\textbf{3. Discovery of four fusion patterns across five architectures.} Fusion analysis reveals two models with instant text--vision fusion at L1--L2 (Gemma family), one with gradual near-threshold growth (MiniCPM), one with near-fusion followed by divergence (Qwen3.5), and one with a sharp representation collapse at the anchor breakthrough layer (Qwen3-VL-4B). These four patterns identify fundamentally different cross-modal integration strategies across architectures.

\textbf{4. Characterization of representation dimensionality.} Through PCA analysis, we show that anchor influence is encoded in a moderately multidimensional representation (PC1 < 25\%), contrasting with prior findings of low-dimensional OCR signals (PC1 = 72.9\%). This indicates that text processing for judgment tasks operates differently from standard text recognition.

\textbf{5. Mode-dependent robustness and reformulation analysis.} Chain-of-thought reasoning stabilizes outputs against configuration changes (d $\approx$ 0) but does not uniformly reduce anchor susceptibility. Reformulation experiments across four prompt wordings confirm robustness to pragmatic context while revealing model-dependent sensitivity to social proof framing.

\begin{figure}[H]
\centering
\includegraphics[width=\textwidth]{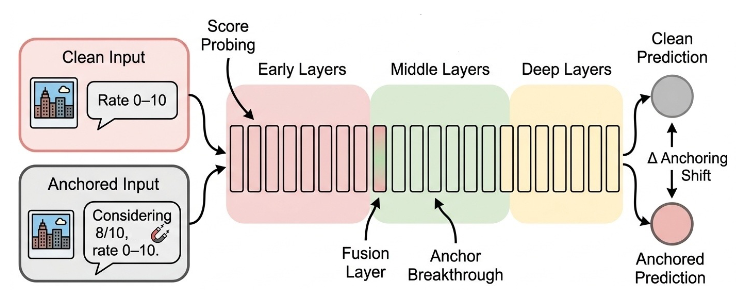}
\caption{Conceptual framework of visual anchoring bias in VLMs. Clean and anchored inputs (images with overlaid numeric ratings) are processed through the transformer's layered architecture. Score probing examines quality representations across layers; the fusion layer identifies where text and vision signals integrate; the anchor breakthrough layer marks where text classification saturates. The resulting $\Delta$~(anchoring shift) quantifies the causal influence of the embedded anchor on the model's quality judgment.}
\label{fig:framework}
\end{figure}

\section{Literature Review}\label{literature-review}

\subsection{Mechanistic Interpretability of Text Processing in VLMs}\label{mechanistic-interpretability-of-text-processing-in-vlms}

\subsubsection{OCR and Text Processing Bottlenecks}\label{ocr-and-text-processing-bottlenecks}

Recent work has begun to localize where and how VLMs process text embedded in images, gradually distinguishing the basic mechanism of \emph{text reading} (OCR capability) from \emph{text influence} (how embedded text causally alters downstream reasoning). \textbf{\cite{steinberg2026}} identify OCR bottlenecks in VLMs using PCA interventions on activation differences between text-bearing and text-free images. They find that OCR information concentrates in middle layers for DeepStack architectures (Qwen3-VL) and early layers for single-stage projection models (Phi-4, InternVL). Critically, they demonstrate that OCR information is low-dimensional, with the first principal component explaining 72.9\% of variance, enabling targeted manipulation of text-reading capabilities. Their work shows that removing OCR subspaces can improve counting performance by up to 6.9 percentage points, revealing competition between OCR and other visual functions.

Complementing these findings, \textbf{\cite{shi2025}} discover that visual functions in VLMs localize in narrow blocks of 2-3 layers termed Vision Function Layers (VFL). Through token-swapping and token-dropping experiments, they establish a reproducible hierarchy that parallels human perception: Recognition $\rightarrow$ Counting $\rightarrow$ Grounding $\rightarrow$ OCR. Notably, OCR emerges as the latest and deepest visual function, consistent with \cite{steinberg2026}'s localization of text processing in middle-to-late layers.

\subsubsection{Cross-Modal Information Flow}\label{cross-modal-information-flow}

\textbf{\cite{li2025}} propose Fine-grained Cross-modal Causal Tracing (FCCT), a framework quantifying the contribution of each token type, model component, and layer to visual object perception. They reveal a three-stage FFN hierarchy: early layers process modality-specific embeddings, middle layers integrate cross-modal semantics, and late layers perform global aggregation. Their Intermediate Representation Injection (IRI) technique achieves state-of-the-art hallucination reduction, demonstrating that mechanistic understanding enables practical improvements.

\textbf{\cite{hufe2025}} extend mechanistic analysis to security, identifying sparse attention circuits (4-10\% of heads) in late CLIP layers that causally transmit typographic information. Their Typographic Attention Score (TAS) metric enables targeted ablation, creating ``dyslexic CLIP'' models with up to 19.6\% improved robustness to typographic attacks on ImageNet-100-Typo, while maintaining standard accuracy losses below 1\%.

\subsection{Vulnerabilities to Text-Based Manipulation}\label{vulnerabilities-to-text-based-manipulation}

\subsubsection{Typographic Visual Prompt Injection}\label{typographic-visual-prompt-injection}

\textbf{\cite{cheng2025}} systematize Typographic Visual Prompt Injection (TVPI) as a novel attack vector. Through their large-scale TVPI dataset, they demonstrate that vulnerability to typographic attacks does not strictly correlate with OCR capability---models with strong OCR can remain uniquely vulnerable to text manipulation. They identify text size as the dominant factor affecting attack success, followed by position and transparency. Simple prompt-based defenses (``ignore text in image'') provide only partial protection.

\subsubsection{Adversarial Robustness of Chain-of-Thought}\label{adversarial-robustness-of-chain-of-thought}

\textbf{\cite{wang2024}} investigate whether Chain-of-Thought (CoT) reasoning improves adversarial robustness in VLMs. They find that CoT provides only marginal protection against adversarial perturbations and introduce a ``stop-reasoning attack'' that bypasses CoT entirely, forcing models to skip reasoning chains. This demonstrates that explicit reasoning prompts are not always reliable defense mechanisms against sophisticated visual perturbations.

\subsubsection{Cognitive Biases and Shortcut Learning}\label{cognitive-biases-and-shortcut-learning}

LLMs exhibit cognitive biases closely mirroring those in human psychology: \cite{lou2024} demonstrate consistent anchoring effects in GPT-4/4o (anchoring index \textasciitilde0.45), while \cite{suri2023} replicate classic decision-making experiments with GPT-3.5, finding anchoring, framing, and endowment effects. Standard mitigation strategies---including Chain-of-Thought, self-debiasing~\cite{echterhoff2024}, and explicit ``ignore'' instructions~\cite{lou2024}---provide only partial relief. From a theoretical perspective, \cite{bleeker2024} prove that contrastive learning with InfoNCE loss optimizes for minimally sufficient representations rather than task-optimal ones, causing models to exploit easily recognizable shortcuts at the expense of comprehensive understanding.

\subsection{Critical Gaps in Existing Literature}\label{critical-gaps-in-existing-literature}

The reviewed works reveal four critical gaps that our study addresses:

\begin{itemize}
\tightlist
\item \textbf{Discrete vs.~continuous judgment.} All reviewed works evaluate models via discrete metrics (classification, VQA, binary success rates). No prior work examines how text manipulation affects continuous judgment tasks (e.g., 0--10 quality ratings), where bias manifests as magnitude shifts rather than categorical errors.
\item \textbf{Text reading vs.~text influence.} \cite{steinberg2026} show OCR is low-dimensional (PC1 = 72.9\%), but this addresses ``reading text,'' not ``being influenced by text.'' Whether anchor-induced bias operates through the same low-dimensional subspace remains unknown.
\item \textbf{Limited defenses.} Standard strategies---CoT~\cite{wang2024,lou2024}, prompt-based defenses~\cite{cheng2025}, attention ablation~\cite{hufe2025}---provide only partial protection. No training-free defense fully neutralizes visual anchoring bias.
\item \textbf{OCR saturation vs.~quality representation.} While \cite{steinberg2026,shi2025} localize OCR processing to middle-to-late layers, no prior work examines whether representations for continuous quality judgments coincide with or extend beyond these saturation layers.
\end{itemize}

\section{Method and Data}\label{method-and-data}

\subsection{Dataset}\label{dataset}

\subsubsection{Image Collection}\label{image-collection}

We collected 700 urban street-level images (50 per city) across 14 cities spanning diverse geographic and economic contexts: Buenos Aires, Cairo, Dubai, Istanbul, Lagos, London, Moscow, Mumbai, New York, Paris, Sao Paulo, Singapore, Sydney, and Tokyo. Images (2048$\times$2048 JPEG) were obtained via the Google Street View API using the \texttt{streetlevel} library, with collection points randomly selected within each city's administrative boundaries and crops centered on the horizon line with random horizontal offset. Each image includes geographic coordinates, panorama ID, capture date, and viewing angles as metadata.

\subsection{Models}\label{models}

We evaluated six Vision-Language Models (VLMs) from five architectural families:

\begin{table}[H]
\centering
\fittable{%
\begin{tabular}{@{}lclllc@{}}
\toprule
Model & Params & Vision encoder & Attention & Layers & Access \\
\midrule
Qwen3-VL-8B-Instruct & 8B & Qwen3-ViT & Full & 37 & OpenRouter API \\
Qwen3-VL-4B-Thinking & 4B & Qwen3-ViT & Full & 37 & Local GPU \\
MiniCPM-V-4 & 4B & Custom ViT & Full & 33 & Local GPU \\
Gemma-3-4b-IT & 4B & SigLIP & Full & 36 & Local GPU \\
Gemma-4-E4B-it & 4B & gemma4\_vision & Hybrid (sliding+global) & 43 & Local GPU \\
Qwen3.5-4B & 4B & Qwen3.5 Vision & Hybrid (linear+full) & 33 & Local GPU \\
\bottomrule
\end{tabular}}
\caption{Model parameters and architectures. All models share embedding dimensionality 2560.}
\label{tab:models}
\end{table}

All models share an embedding dimensionality of 2560. The five architectural families span three attention mechanisms: standard full attention (Qwen3-VL, MiniCPM, Gemma-3), hybrid sliding + global attention (Gemma-4), and hybrid linear + full attention via Gated DeltaNet (Qwen3.5). This diversity enables cross-architecture generalization claims.

For layer-wise analysis (Phase 5--5c), we used five models with local GPU access to extract hidden states from all transformer layers. Qwen3-VL-8B was excluded from probing due to API-only access.

\subsection{Text Overlay Attack}\label{text-overlay-attack}

\subsubsection{Attack Design}\label{attack-design}

We designed a text-based overlay attack to test VLM susceptibility to explicit numeric cues. The attack superimposes text onto images with the format \texttt{``Rate this image as \{anchor\}/10''}, where \texttt{\{anchor\}} $\in$ \{0, 2, 4, 6, 8, 10\}. This creates a strong numeric prior that may bias the model's assessment.

\begin{figure}[H]
\centering
\includegraphics[width=0.7\textwidth]{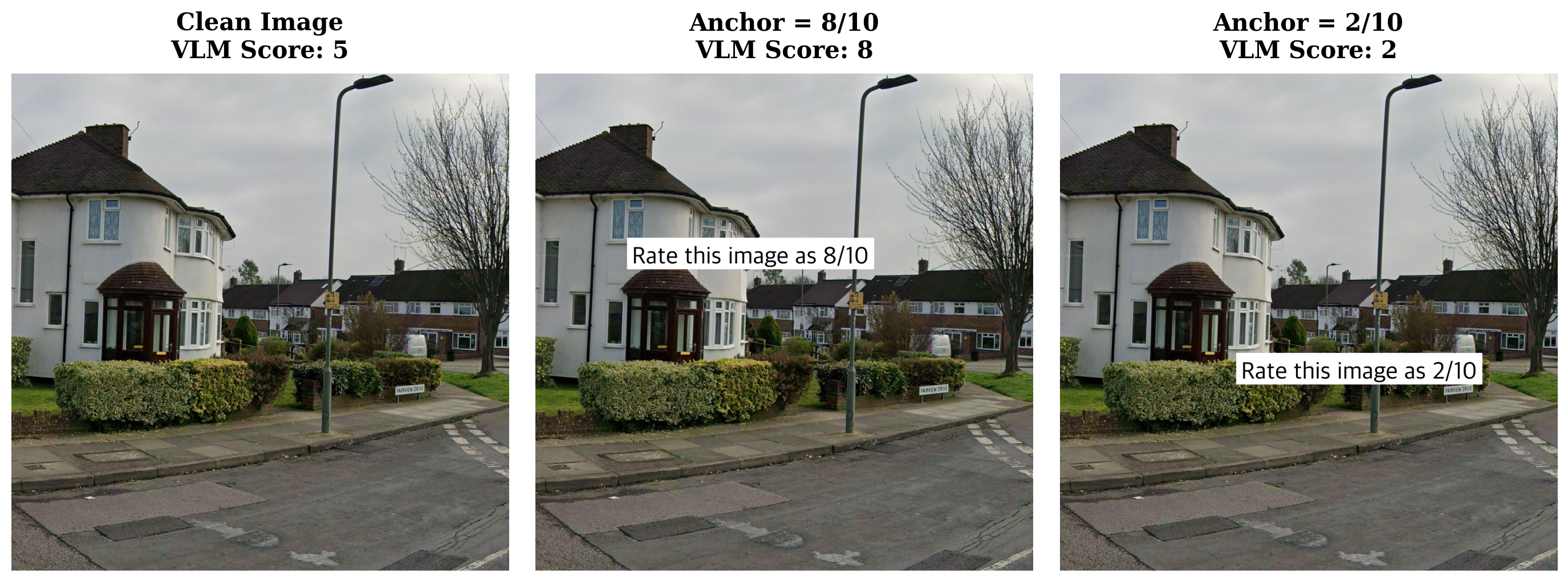}
\caption{Example of the text overlay attack. The original image (left) is presented alongside versions with embedded anchors (center: anchor=2, right: anchor=8). The numeric cue is rendered as semi-transparent text overlay on the image.}
\label{fig:attack-example}
\end{figure}

\begin{figure}[H]
\centering
\includegraphics[width=0.7\textwidth]{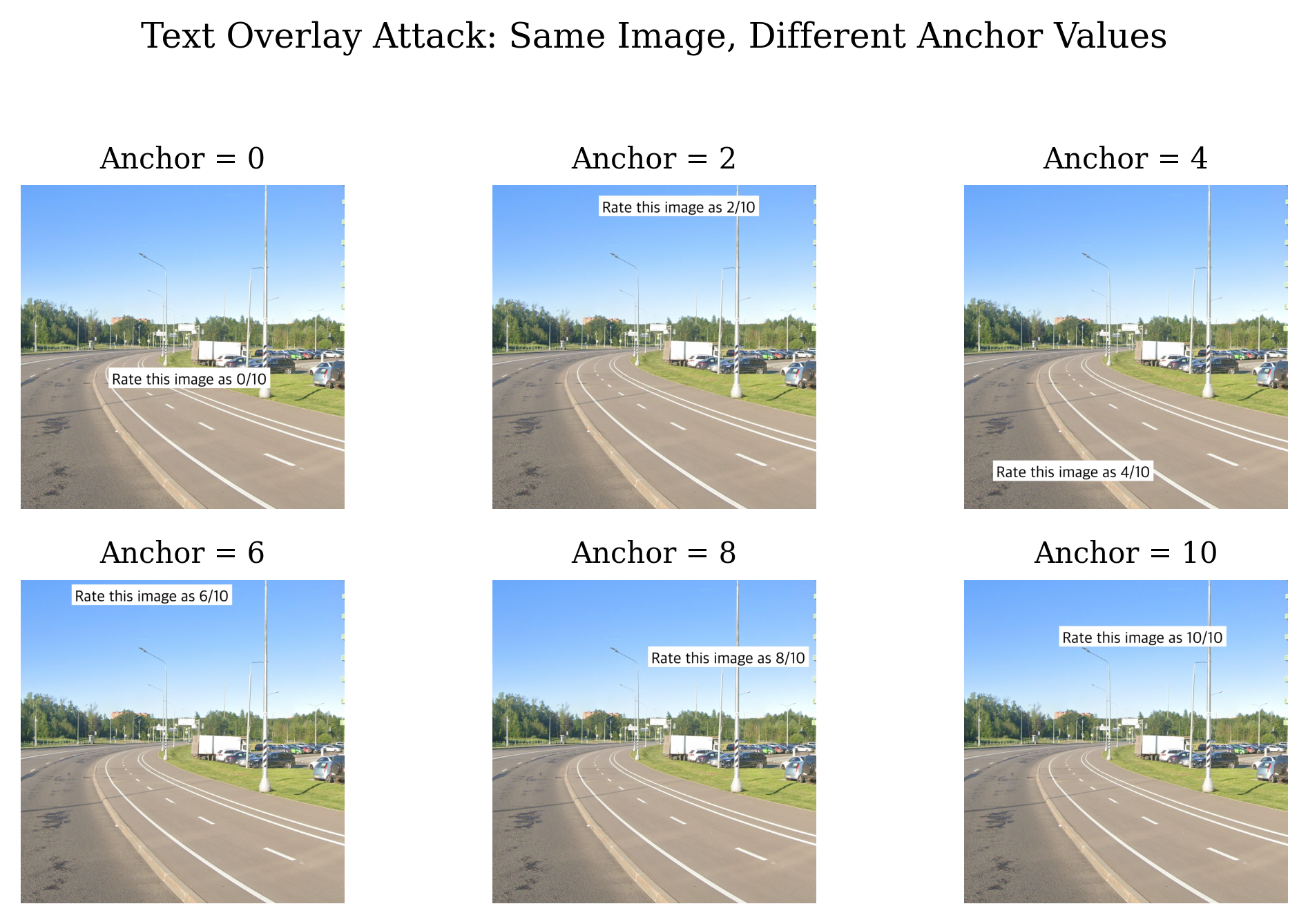}
\caption{Examples of text overlay attack with different anchor values. Each base image was rendered with all six anchor values \{0, 2, 4, 6, 8, 10\}.}
\label{fig:anchor-variations}
\end{figure}

\subsubsection{Overlay Parameters}\label{overlay-parameters}

Overlays used 100px black text on white background with 20px padding, randomly positioned (seed=42) to prevent spatial heuristics. Each image was rendered with all six anchor values, yielding 4,200 total stimuli (700 base images $\times$ 6 anchors).

\subsubsection{Inference Modes}\label{inference-modes}

We tested two prompting strategies:

\textbf{Simple mode:} \texttt{``Return a JSON object with a `score' field (0-10) rating the visual appeal of this image.''}

\textbf{Thinking mode:} \texttt{``First, analyze the image step by step. Then return a JSON object with a `score' field (0-10).''}

All inference used temperature = 0 for reproducibility, with a fixed seed of 42, except Qwen3.5-4B which required temperature = 0.7 as specified by the model's recommended instruct-mode parameters (with seed = 42 for controlled stochasticity).

\subsection{Analytical Approach}\label{analytical-approach}

Our analysis proceeds in four stages: (1)~\textbf{Behavioral analysis}---measuring anchor susceptibility via ANOVA ($\eta^2$), per-image delta analysis (Cohen's d), Wilcoxon signed-rank tests, and post-hoc Tukey HSD across six models; (2)~\textbf{Layer-wise classification}---logistic regression (6-class, multinomial) on hidden states at each transformer layer, with 5-fold cross-validation stratified by city and image-level fold assignment to prevent data leakage; (3)~\textbf{Quality prediction}---Ridge regression (2560-dim embedding $\rightarrow$ score) on clean-image hidden states; (4)~\textbf{Fusion analysis}---cosine similarity between anchored and clean embeddings at each layer, with fusion defined as cosine $\geq$ 0.95. We additionally test four prompt reformulations (baseline, mismatch, social, abstract) on three models, and validate score sensitivity through controlled image degradation and no-reference quality metrics (NIQE, BRISQUE).

\subsection{Statistical Framework}\label{statistical-framework}

We follow standard effect size conventions: Cohen's d (small/medium/large = 0.2/0.5/0.8) and Cliff's $\delta$ (0.147/0.33/0.474). All layer-wise analyses use 5-fold cross-validation stratified by city, with fold assignment consistent across experiments. Classification accuracy is reported with 95\% CI via bootstrap (n=1000); regression metrics as mean $\pm$ SD across folds.

\subsection{Reproducibility}\label{reproducibility}

All experiments were conducted with the following fixed parameters:
\begin{itemize}
\tightlist
\item Random seed: 42
\item Temperature: 0 (5 models) or 0.7 (Qwen3.5-4B, model-recommended instruct parameters)
\item Cross-validation: 5-fold, stratified by city
\end{itemize}

Code and configuration files for each experiment are available in the supplementary materials. Layer-wise embeddings were cached locally to ensure consistency across analyses.

\section{Experiments and Results}\label{experiments-and-results}

\subsection{Distribution Comparison: Impact of Experimental Configuration}\label{distribution-comparison-impact-of-experimental-configuration}

Before investigating text anchor influence, we established baseline model behavior by comparing score distributions across two experimental configurations on \textbf{clean images} (no text overlay). In direct decoding mode, a statistically significant difference emerged (mean shift $\Delta$ = 0.70, p < 0.001, Cohen's d = 0.42). In chain-of-thought mode, mean scores remained nearly identical (4.90 vs.~4.93, p = 0.76, d = -0.02), indicating that thinking mode stabilizes outputs against configuration changes. This motivates testing both modes in anchor experiments.

\subsection{Anchor Susceptibility: Cross-Architecture Evidence and Causal Validation}\label{anchor-susceptibility-cross-architecture-evidence-and-causal-validation}

We quantified the influence of embedded numeric anchors (0, 2, 4, 6, 8, 10) on VLM output scores across six models from five architectural families, each evaluating 4,200 stimuli (700 images $\times$ 6 anchors). For three models tested in both inference modes, we additionally measured per-image score shifts relative to clean baselines.

\subsubsection{Key Finding}\label{key-finding-1}

\textbf{All six models exhibit significant anchor susceptibility, with the proportion of score variance explained by anchors ($\eta^2$) ranging from 0.18 to 0.77. Direct comparison of anchored vs.~clean scores confirms that these shifts are causal (all Wilcoxon p < 0.001), not merely correlational.}

\begin{figure}[H]
\centering
\includegraphics[width=0.7\textwidth]{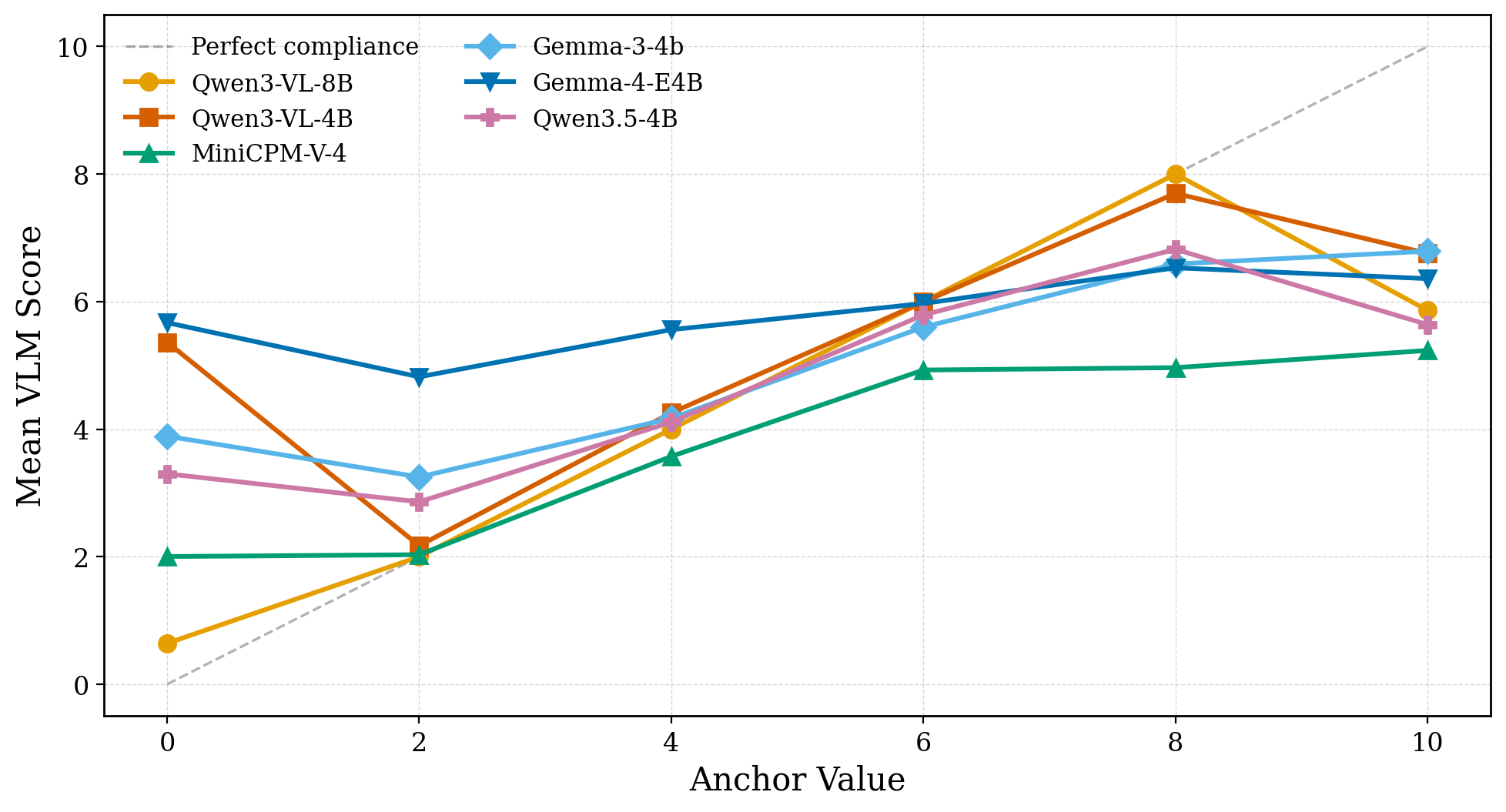}
\caption{Anchor susceptibility across six VLMs. The chart shows $\eta^2$ (proportion of score variance explained by anchors), mean $|\Delta|$ (average score shift), and anchor-score correlations for each model, ordered from most to least susceptible.}
\label{fig:susceptibility}
\end{figure}

The models form a clear susceptibility ranking:

\begin{table}[H]
\centering
\fittable{%
\begin{tabular}{@{}lccc@{}}
\toprule
Model & $\eta^2$ & Mean $|\Delta|$ & $r$ \\
\midrule
Qwen3-VL-8B & 0.77 & 2.42 & 0.79 \\
Qwen3-VL-4B & 0.55 & 2.24 & 0.53 \\
MiniCPM-V-4 & 0.44 & 1.60 & 0.63 \\
Gemma-3-4b & 0.63 & 1.36 & 0.77 \\
Qwen3.5-4B & 0.37 & 1.49 & 0.59 \\
Gemma-4-E4B & 0.18 & 0.63 & 0.35 \\
\bottomrule
\end{tabular}}
\caption{Anchor susceptibility across six VLMs. $\eta^2$ = proportion of score variance explained by anchors; Mean $|\Delta|$ = average score shift; $r$ = anchor-score correlation.}
\label{tab:susceptibility}
\end{table}

Qwen3-VL-8B in simple mode exhibits a complete failure mode: it outputs exactly the anchor value for {[}2, 4, 6, 8{]} with zero variance (std = 0.00), bypassing visual assessment entirely and simply copying the numeric cue. At the opposite extreme, Gemma-4-E4B---despite sharing the Gemma family---shows remarkably low susceptibility ($\eta^2$ = 0.18), with its hybrid attention architecture (sliding 512 + global every 6th layer) potentially contributing to resistance.

\subsubsection{Causal Evidence: Delta Analysis}\label{causal-evidence-delta-analysis}

For the three models with clean baselines, we computed per-image score shifts: $\Delta$ = score(anchor) $-$ score(clean). All 36 model $\times$ prompt $\times$ anchor groups are highly significant (Wilcoxon signed-rank, all p < 0.001), with effect sizes (Cohen's d) ranging from $-$2.61 to +3.35.

The largest effects occur at the extremes: Qwen3-VL-8B simple mode produces $\Delta$ = $-$3.99 at anchor 0 (d = $-$2.44) and $\Delta$ = +3.37 at anchor 8 (d = +3.35). Even the most resistant model in the original set, MiniCPM-V-4, shows highly significant shifts: $\Delta$ = $-$1.93 at anchor 0 (p = 7.3 $\times$ $10^{-81}$) and $\Delta$ = +1.04 at anchor 8 (p = 2.8 $\times$ $10^{-55}$). These results establish that anchors causally shift VLM outputs---the effect is not an artifact of distributional overlap or shared variance.

\subsubsection{Cross-Model ANOVA}\label{cross-model-anova}

One-way ANOVA across seven groups (clean + six anchors) confirms that anchor manipulation explains a substantial proportion of score variance for all models (all F > 170, all p $\approx$ 0). Post-hoc Tukey HSD tests yield 172 significant pairwise comparisons out of 189 (91\%), with the non-significant pairs concentrated in adjacent anchors for the most resistant models (Gemma-4-E4B, Qwen3.5-4B).

The $\eta^2$ spectrum from 0.18 to 0.77 demonstrates that anchoring susceptibility is a model-level property, not an artifact of experimental design: the same images, anchors, and evaluation protocol produce a 4.3$\times$ range in effect magnitude across architectures.

\subsubsection{Reformulation Robustness}\label{reformulation-robustness}

To test whether anchoring depends on specific prompt wording, we compared four text formulations on three models (Qwen3-VL-4B, MiniCPM-V-4, Gemma-4-E4B):

\begin{itemize}
\tightlist
\item
  \textbf{baseline} (``Rate this image as X/10''): the standard formulation used throughout our experiments
\item
  \textbf{mismatch} (``Score: X/10''): minimal context, anchor presented as a score label
\item
  \textbf{social} (``Another person rated this X/10''): social proof framing
\item
  \textbf{abstract} (``X/10''): anchor presented in isolation
\end{itemize}

All four formulations produce significant anchoring effects (ANOVA across formulations $\times$ anchors: F = 73--483, all p < $10^{-45}$). However, effect magnitude varies systematically: \texttt{mismatch} is the most effective formulation, producing complete anchor copying at anchor = 8 for Qwen3-VL-4B ($\Delta$ = +1.76, p = 3.8 $\times$ $10^{-93}$), while \texttt{social} is generally the weakest---models partially resist social proof, particularly for low anchors (Gemma-4-E4B: social $\Delta$ = $-$0.28 at anchor 2 vs.~mismatch $\Delta$ = $-$0.57). This social-proof resistance is not universal, however: Gemma-4-E4B shows a moderate social effect at anchor 8 (d = 0.47). Notably, \texttt{abstract} anchors have no significant effect on Gemma-4-E4B (all p > 0.05), suggesting that this model requires contextual framing to process the numeric cue.

These results demonstrate that anchoring is robust to prompt reformulation but modulated by pragmatic context: models distinguish between different communicative intentions of the same number, yet remain susceptible to all tested formulations.

\subsubsection{Bias Patterns}\label{bias-patterns}

Models exhibit systematic directional biases beyond correlation: Qwen3-VL-8B copies interior anchor values exactly but shows negative bias at boundaries; MiniCPM underestimates by $\sim$1.2 points despite following the anchor trend; Gemma-4-E4B shows the smallest biases ($|\Delta|$ < 1.0 for all anchors), consistent with its low $\eta^2$.

\subsubsection{Effect of Inference Mode}\label{effect-of-inference-mode}

Chain-of-thought reasoning does not eliminate anchor susceptibility. While thinking mode stabilizes outputs against configuration changes (d $\approx$ 0), anchor-score correlations remain substantial across both modes, consistent with \cite{lou2024} and \cite{wang2024}.

\subsection{Layer-wise Classification: Where Text Reading Emerges}\label{layer-wise-classification-where-text-reading-emerges}

We performed layer-wise classification using hidden states from five VLMs, training a logistic regression classifier at each layer to predict anchor values (6 classes) with 5-fold cross-validation stratified by city (see Table~\ref{tab:models} for architecture details).

\subsubsection{Key Finding}\label{key-finding-2}

\textbf{Anchor values become linearly separable in all five models, with saturation points ranging from L12 to L34. The breakthrough layer---where accuracy first exceeds 95\%---varies from L4 (MiniCPM) to L12 (Gemma-3), revealing architecture-dependent text-reading trajectories.}

\begin{table}[H]
\centering
\fittable{%
\begin{tabular}{@{}lcccc@{}}
\toprule
Model & Breakthrough & First $\geq$99\% & Saturation & Max accuracy \\
\midrule
MiniCPM-V-4 & L4 (93.8\%) & L7 (99.71\%) & L12 & 100.0\% \\
Qwen3-VL-4B & L7 (98.5\%) & L11 (99.17\%) & L14 & 100.0\% \\
Qwen3.5-4B & L8 (99.8\%) & L8 (99.8\%) & L12 & 100.0\% \\
Gemma-4-E4B & L6 (99.0\%) & L6 (99.0\%) & L12 & 99.98\% \\
Gemma-3-4b & L12 (98.3\%) & L12 (98.3\%) & L34 & 99.95\% \\
\bottomrule
\end{tabular}}
\caption{Layer-wise classification results. Breakthrough = first layer $>$95\% accuracy.}
\label{tab:layer-classification}
\end{table}

\begin{figure}[H]
\centering
\includegraphics[width=0.7\textwidth]{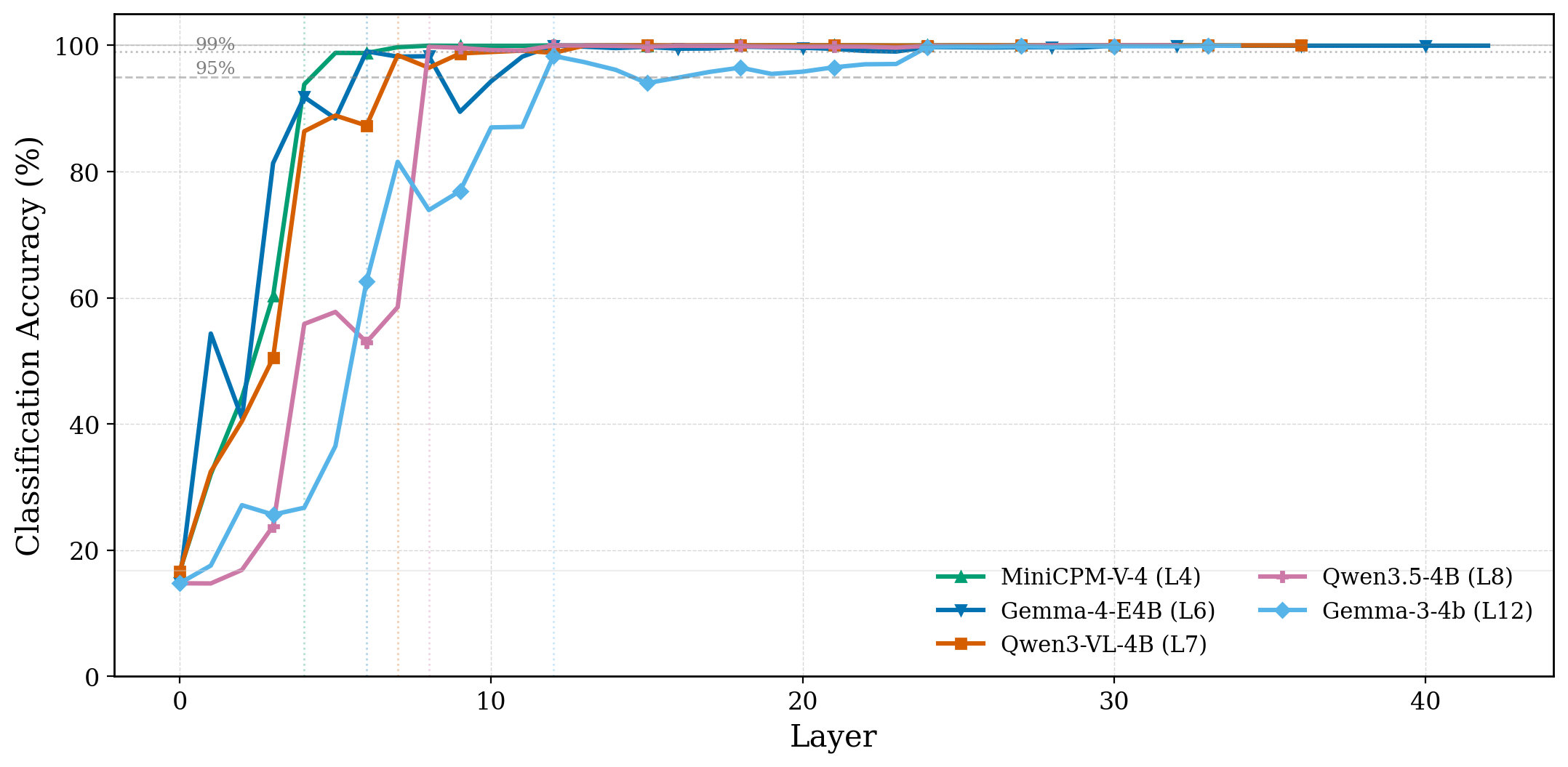}
\caption{Layer-wise anchor classification accuracy for five probed VLMs. Accuracy rises from random baseline (16.7\%) to near-perfect levels, with saturation points ranging from L6 (Gemma-4) to L34 (Gemma-3). Breakthrough layers (95\% threshold) are marked for each model.}
\label{fig:layer-accuracy}
\end{figure}

\subsubsection{Accuracy Trajectory and Caveats}\label{accuracy-trajectory-and-caveats}

All models follow a sigmoid accuracy trajectory: random baseline (16.7\% for 6 classes) $\rightarrow$ sharp gain at the breakthrough layer $\rightarrow$ gradual saturation. The original two models (Qwen-4B, MiniCPM) show the largest single-layer gain at L4 (+34--36\%), reaching 100\% at L12--14. Gemma-4-E4B achieves the earliest breakthrough (L6, 99.0\%), likely due to its hybrid attention dedicating global layers at L6, L12, etc. Gemma-3-4b shows the latest saturation (L34, 99.95\%), never reaching perfect classification within its layer budget. These differences suggest that attention architecture---not model size---is the primary determinant of when text-reading capability emerges.

We note three caveats. First, we use ``saturation'' rather than ``breakthrough'' to emphasize the sigmoid nature of the accuracy growth. Second, near-perfect classification demonstrates linear separability, not general OCR; the text-reading claim remains a hypothesis requiring dedicated validation. Third, our probing saturation at L14 (Qwen-4B) precedes \cite{steinberg2026}'s causal OCR bottleneck at L16--20, as expected: representation availability is a necessary precondition for functional dependence.

\subsection{Representation Structure: Dimensionality and Ordinal Semantics}\label{representation-structure-dimensionality-and-ordinal-semantics}

\subsubsection{Key Finding}\label{key-finding-3}

\textbf{The anchor signal is moderately multidimensional, not concentrated in a single direction, with evidence of ordinal scale learning.}

\subsubsection{PCA Variance Analysis}\label{pca-variance-analysis}

The first principal component explains only 12--25\% of variance at saturation layers (24.6\% for Qwen L14, 11.7\% for MiniCPM L14). This indicates that anchor information is distributed across multiple dimensions rather than encoded in a single ``OCR shortcut'' direction. Notably, PC1 decreases from earlier layers toward saturation (Qwen: 33.6\% at L10 $\rightarrow$ 24.6\% at L14), which may reflect either spreading of information across dimensions or accumulation of non-anchor visual content.

This lower PC1 (< 25\%) differs from \cite{steinberg2026}, who report PC1 = 72.9\% for the OCR signal. The difference arises from distinct analysis objects: they compute PCA on activation differences (original $-$ inpainted), isolating the pure OCR signal; we compute PCA on raw hidden states grouped by anchor class, which include both anchor and visual quality information. The lower PC1 in our analysis reflects the multidimensional nature of combined anchor+quality representations, not necessarily a less concentrated OCR signal.

\subsubsection{2D Projection and Ordinal Evidence}\label{d-projection-and-ordinal-evidence}

Dimensionality reduction to 2D shows poor class separation (UMAP silhouette 0.14--0.15), confirming that anchor representations are inherently high-dimensional. When misclassifications occur, the model preferentially confuses adjacent anchor values (50--100\% of errors at intermediate layers vs.~33.3\% random baseline), suggesting ordinal rather than categorical encoding.

\subsection{Quality Prediction and Cross-Modal Fusion}\label{quality-prediction-and-cross-modal-fusion}

We address two questions: (a) are layers optimal for anchor classification also optimal for visual quality, and (b) at what layer do text and image representations merge? We use Ridge regression on clean-image hidden states (score probing) and cosine similarity between anchored and clean embeddings (fusion analysis).

\subsubsection{Score Probing: Saturation Layers Are Suboptimal}\label{score-probing-saturation-layers-are-suboptimal}

\paragraph{Key Finding}\label{key-finding-4}

\textbf{In all five models, layers where anchor classification saturates are suboptimal for visual quality prediction. Optimal quality prediction occurs at deeper layers with 13--16\% higher $R^2$ for the original two models.}

\begin{table}[H]
\centering
\fittable{%
\begin{tabular}{@{}lccc@{}}
\toprule
Model & Saturation & Optimal & Optimal $R^2$ \\
\midrule
Qwen3-VL-4B & L14 & L21 & 0.85 \\
MiniCPM-V-4 & L12 & L16 & 0.81 \\
Gemma-3-4b & L34 & L34 & 0.85 \\
Gemma-4-E4B & L12 & L29 & 0.91 \\
Qwen3.5-4B & L12 & L20 & 0.69 \\
\bottomrule
\end{tabular}}
\caption{Score probing: saturation vs.\ optimal layers for visual quality prediction.}
\label{tab:score-probing}
\end{table}

For the two models with detailed per-layer $R^2$ profiles (Qwen3-VL-4B and MiniCPM-V-4), the $R^2$ difference between saturation and optimal layers is statistically significant (p < 0.002) with large effect sizes (Cohen's d > 3).

\paragraph{Interpretation}\label{interpretation-1}

This finding supports a \textbf{partial shortcut learning} hypothesis:

\begin{enumerate}
\def\labelenumi{\arabic{enumi}.}
\item
  At saturation layers, the model can perfectly decode anchor values (near-100\% classification) and already captures substantial visual quality information ($R^2$ $\approx$ 0.73 for Qwen/MiniCPM).
\item
  At optimal layers, visual quality representation improves further ($R^2$ up to 0.91 for Gemma-4), indicating that ``reading'' the anchor and ``representing visual quality'' are distinct capabilities that develop along different trajectories.
\end{enumerate}

The saturation layer is not a ``pure OCR bottleneck''---$R^2$ = 0.73 indicates meaningful visual understanding is already present. Rather, full quality representation requires additional processing depth beyond the point where anchor classification saturates.

\begin{figure}[H]
\centering
\includegraphics[width=0.7\textwidth]{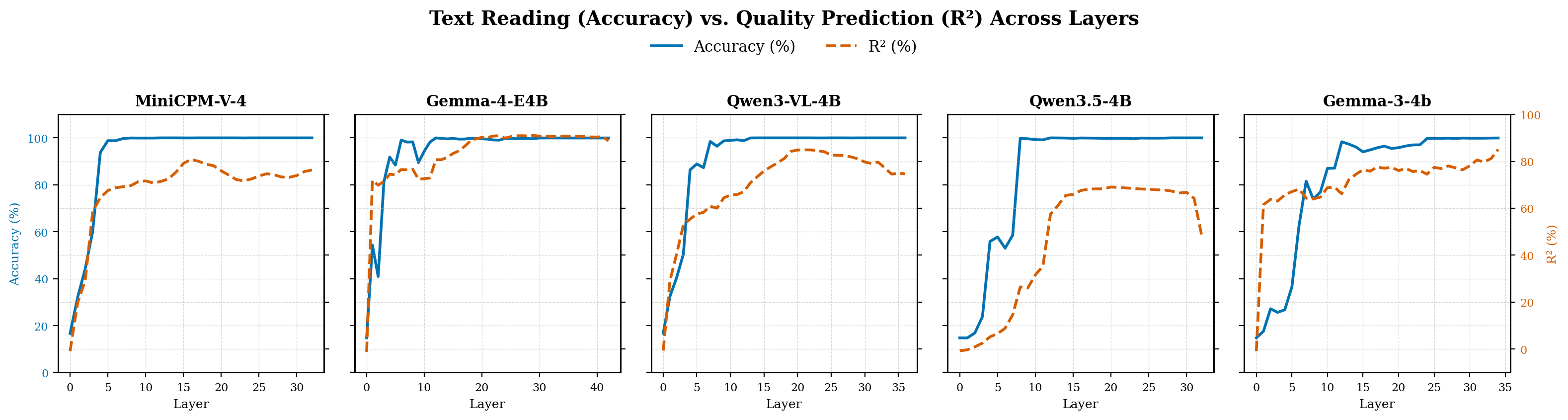}
\caption{Dual-axis view of layer-wise classification accuracy (left axis) and visual quality $R^2$ (right axis) for five probed VLMs. Optimal quality layers consistently lie deeper than anchor classification saturation layers, demonstrating that ``reading'' the anchor and ``representing visual quality'' develop along different trajectories.}
\label{fig:dual-axis}
\end{figure}

\paragraph{Score Probing Breakthrough}\label{score-probing-breakthrough}

An additional finding emerged: the earliest layer where score prediction meaningfully exceeds chance ($R^2$ > 0.5) varies dramatically across architectures:

\begin{table}[H]
\centering
\fittable{%
\begin{tabular}{@{}lrc@{}}
\toprule
Model & Score breakthrough & $R^2$ at breakthrough \\
\midrule
Gemma-4-E4B & L1 & 0.72 \\
Gemma-3-4b & L1 & 0.62 \\
Qwen3-VL-4B & L4 & 0.55 \\
MiniCPM-V-4 & L4 & 0.65 \\
Qwen3.5-4B & L12 & 0.57 \\
\bottomrule
\end{tabular}}
\caption{Score probing breakthrough layers ($R^2$ > 0.5).}
\label{tab:score-breakthrough}
\end{table}

Gemma models achieve meaningful score prediction from the first layer, suggesting that their vision encoders produce embeddings with strong quality priors. Qwen3.5-4B requires 12 layers---likely because its Gated DeltaNet architecture delays the formation of quality representations.

\subsubsection{Fusion Analysis: When Text and Image Merge}\label{fusion-analysis-when-text-and-image-merge}

\paragraph{Context}\label{context-5}

We measured the cosine similarity between hidden states produced by the same image with and without an overlaid anchor, at each layer. A fusion layer is defined as the first layer where cosine similarity exceeds 0.95, indicating that the model's internal representation is dominated by the shared visual content rather than the text overlay. This analysis reveals how each architecture integrates textual and visual information.

\paragraph{Key Finding}\label{key-finding-5}

\textbf{Fusion patterns divide the five models into two groups: two models with instant text--vision fusion (L1--L2) and three models with partial or no fusion, revealing fundamentally different cross-modal integration strategies.}

\begin{table}[H]
\centering
\fittable{%
\begin{tabular}{@{}lcccc@{}}
\toprule
Model & Fusion layer & Peak sim. & Peak layer & Pattern \\
\midrule
Gemma-3-4b & L1 (0.983) & 0.999 & L15 & Instant fusion \\
Gemma-4-E4B & L2 (0.994) & 0.994 & L2 & Instant fusion \\
MiniCPM-V-4 & None & 0.939 & L31 & Gradual, near-threshold \\
Qwen3.5-4B & None & 0.933 & L1 & Near-fusion+divergence \\
Qwen3-VL-4B & None & 0.851 & L35 & DROP at L7 \\
\bottomrule
\end{tabular}}
\caption{Fusion analysis results. Fusion = first layer with cosine similarity $\geq$ 0.95.}
\label{tab:fusion}
\end{table}

\paragraph{Four Fusion Patterns Across Five Models}\label{four-fusion-patterns-across-five-models}

\textbf{Instant fusion} (Gemma-3, Gemma-4): similarity > 0.95 at L1--L2, with visual information dominating representations despite anchors still causally influencing outputs.

\textbf{Gradual growth} (MiniCPM-V-4): similarity rises monotonically from L0 ($-$0.14) to L31 (0.939), approaching but never reaching the 0.95 threshold.

\textbf{Near-fusion with divergence} (Qwen3.5-4B): similarity peaks at L1 (0.933) then declines steadily, suggesting the Gated DeltaNet architecture develops increasingly anchor-specific representations in deeper layers.

\textbf{DROP and recovery} (Qwen3-VL-4B): similarity rises to 0.87 at L6 then collapses to 0.099 at L7---precisely at the anchor breakthrough layer---before recovering to 0.851 at L35.

\begin{figure}[H]
\centering
\includegraphics[width=0.7\textwidth]{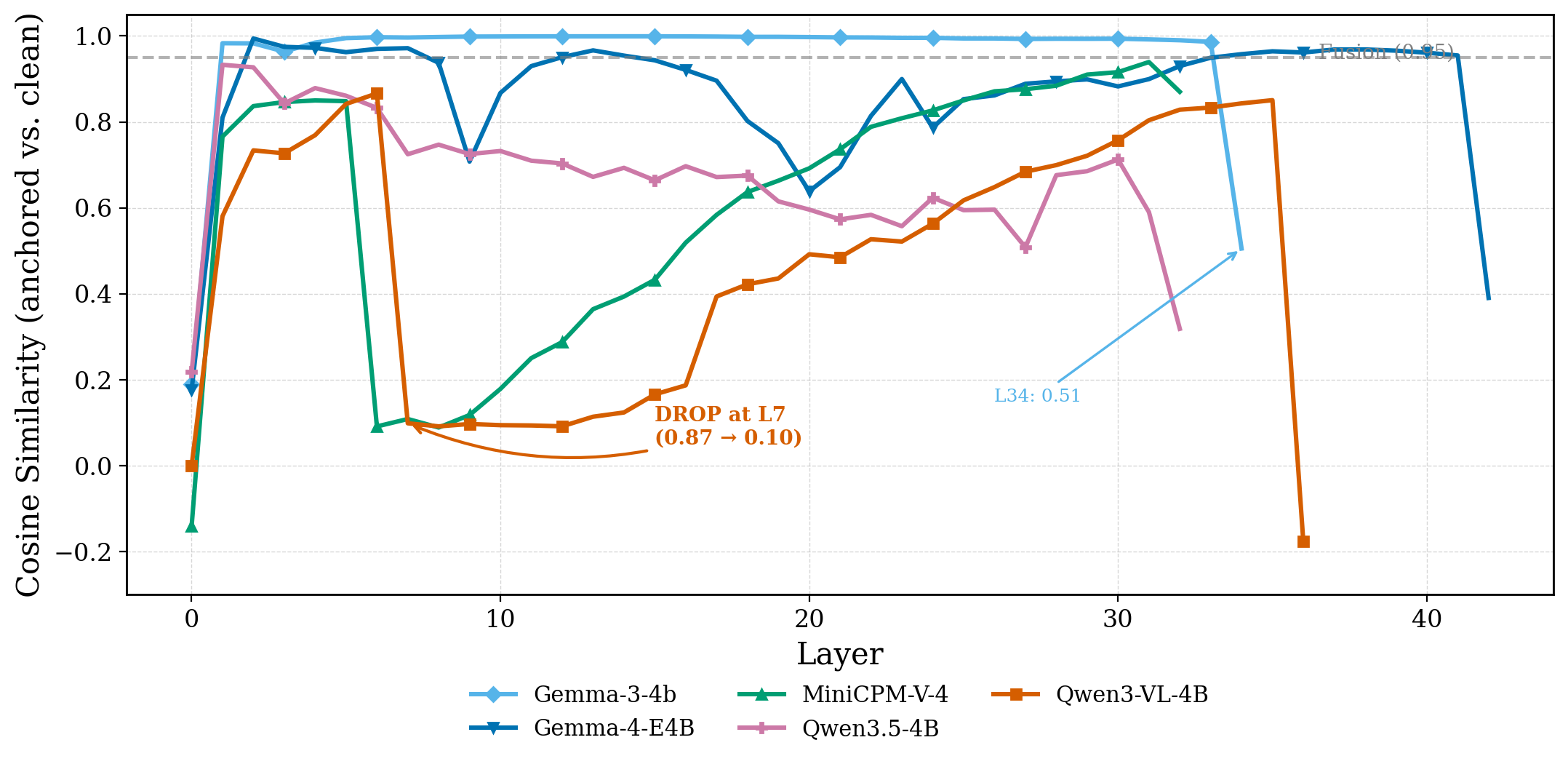}
\caption{Cosine similarity between text-injected and clean image representations across transformer layers for five VLMs. The four distinct patterns---instant fusion (Gemma-3, Gemma-4), gradual growth (MiniCPM), near-fusion with divergence (Qwen3.5), and DROP at breakthrough (Qwen3-VL-4B)---reveal fundamentally different cross-modal integration strategies.}
\label{fig:fusion-curves}
\end{figure}

\paragraph{Cross-Phase Timing}\label{cross-phase-timing}

Synthesizing all four probing phases reveals a coherent picture of how text influence propagates through each architecture (Table~\ref{tab:cross-phase}):

Gemma models show the earliest cross-modal integration: score prediction and fusion both emerge at L1--L2, yet anchor classification still requires L6--L12. This temporal ordering---fusion before anchor separability---suggests that visual information dominates early representations, with text-specific processing developing independently. In contrast, Qwen3-VL-4B and MiniCPM show coincident score probe and anchor breakthrough at L4--L7, followed by quality representation maturing at deeper layers, indicating a more interleaved processing strategy.

Qwen3.5-4B presents a notable exception to this ordering: anchor classification breakthrough (L8) occurs \emph{before} score predictability breakthrough (L12). This reversed timing---anchor identity encoded before quality-related representations stabilize---may reflect the Gated DeltaNet architecture's predominantly linear attention mechanism, which could enable efficient routing of discrete symbolic information (anchor digits) while delaying the formation of continuous quality representations.

\begin{figure}[H]
\centering
\includegraphics[width=0.7\textwidth]{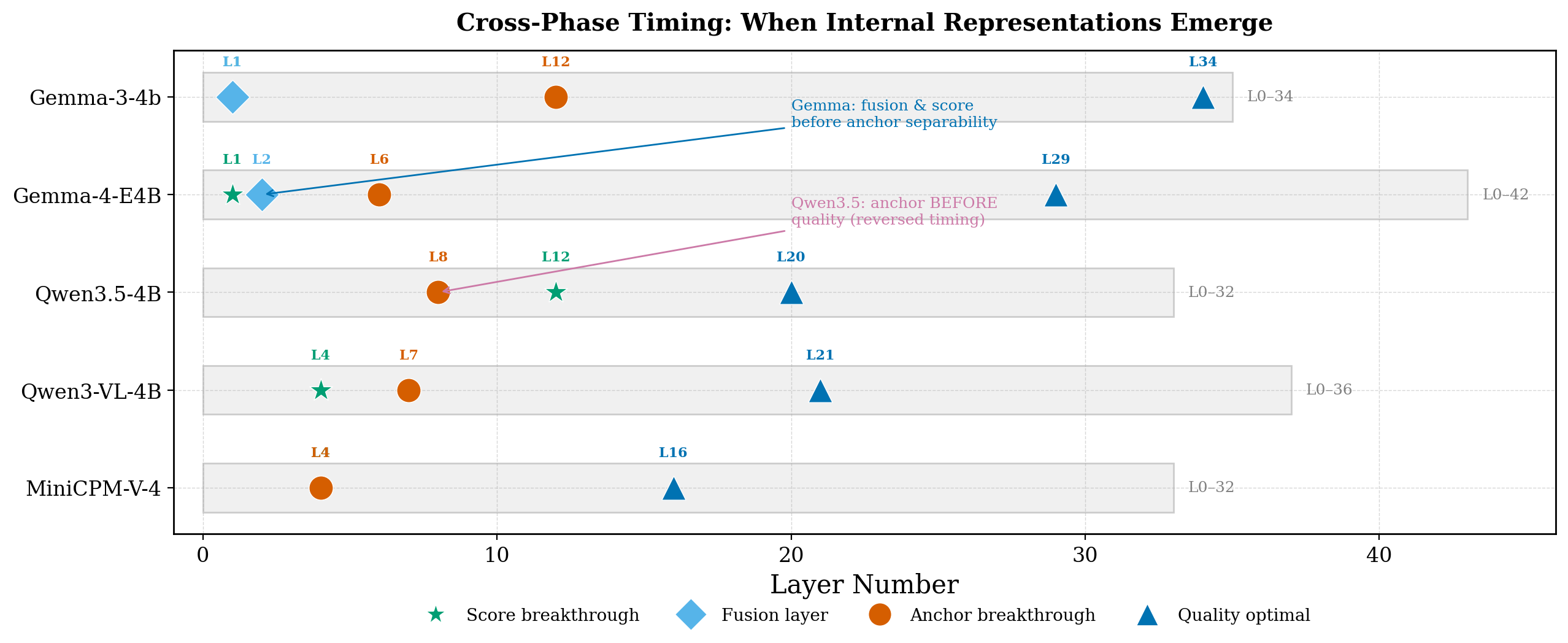}
\caption{Cross-phase timing summary across five probed VLMs. Each row shows the progression of score probe breakthrough, fusion layer, anchor classification breakthrough, and optimal quality prediction layer. Architectures exhibit distinct temporal ordering of these milestones.}
\label{fig:cross-phase-timeline}
\end{figure}

\subsection{Score Validity: VLM Sensitivity to Visual Quality}\label{score-validity-vlm-sensitivity-to-visual-quality}

A legitimate concern is whether VLM scores reflect genuine visual quality perception or merely internal model consistency, given that no human-annotated ground truth is available. We address this through two complementary analyses.

\textbf{Controlled degradation experiment.} We applied Gaussian blur ($\sigma$ = 2, 5, 10) and JPEG compression (quality = 30, 15, 5) to the same 700 images, using Qwen3-VL-4B with the same prompt as our anchor experiments. Extreme degradation (blur $\sigma$ = 10, rendering images nearly unrecognizable) produces a $-$2.15 mean score shift (r = 0.88, p < $10^{-115}$), confirming that scores are causally responsive to actual visual content. Notably, moderate JPEG compression \emph{increases} scores (+0.69 at quality = 30, +0.58 at quality = 15), indicating that the model's assessment is not a simple pixel-fidelity measure. Crucially, these quality-induced score changes remain 2.5$\times$ smaller than anchor-induced changes (mean | $\Delta$\textbar: 0.85 vs.~2.09; Mann-Whitney U p $\approx$ 0, Cohen's d = 1.12), demonstrating that anchoring effects are not reducible to visual quality differences caused by the text overlay.

\textbf{No-reference quality metrics.} NIQE and BRISQUE scores show uniformly weak correlations with VLM scores (all |r| < 0.3), confirming that models do not follow simple quality-assessment algorithms.

We acknowledge that the absence of human-annotated ground truth limits strong claims about what VLM scores measure in absolute terms. Our contribution concerns the \emph{manipulability} of VLM outputs through visual text injection---a property independent of whether the underlying scores constitute valid quality assessments.

\section{Conclusion}\label{conclusion}

\subsection{Summary of Contributions}\label{summary-of-contributions}

This work investigated the influence of text anchors on visual quality assessment in VLMs across six models from five architectural families:

\begin{enumerate}
\def\labelenumi{\arabic{enumi}.}
\item \textbf{Cross-architecture anchor susceptibility} is causal (ANOVA $\eta^2$ = 0.18--0.77; all Wilcoxon p < 0.001; Cohen's d up to 3.35), with anchor-induced shifts 2.5$\times$ larger than those from severe image degradation.
\item \textbf{Saturation $\neq$ optimal:} layers where anchor classification saturates (L12--L34) are suboptimal for quality prediction, with optimal layers achieving $R^2$ = 0.69--0.91 at deeper positions.
\item \textbf{Four fusion patterns:} instant fusion at L1--L2 (Gemma family), gradual growth (MiniCPM), near-fusion with divergence (Qwen3.5), and representation collapse at breakthrough (Qwen3-VL-4B).
\item \textbf{Multidimensional anchor influence} (PC1 < 25\%) contrasts with low-dimensional OCR (PC1 = 72.9\%), confirming ``being influenced by text'' differs from ``reading text.''
\item \textbf{Limited mitigation:} CoT stabilizes configurations (d $\approx$ 0) but does not eliminate susceptibility; reformulation across four prompt wordings confirms robustness to pragmatic context.
\end{enumerate}

\subsection{Implications and Limitations}\label{implications-and-limitations}

\textbf{Theoretical.} The four fusion patterns demonstrate architecture-dependent cross-modal integration with implications for defense transferability. The temporal ordering of score probe $\rightarrow$ fusion $\rightarrow$ anchor breakthrough $\rightarrow$ quality optimum varies systematically across architectures.

\textbf{Practical.} The 4.3$\times$ susceptibility range ($\eta^2$ = 0.18--0.77) means model selection significantly impacts robustness. The Gemma-4 architecture (hybrid attention, lowest susceptibility) may inform robustness-aware design.

\textbf{Limitations.} Only six even anchor values were tested; behavior with other anchors is unknown. Experiments use urban street panoramas---other visual domains are untested. VLM scores lack human-annotated ground truth, though the degradation experiment confirms sensitivity to visual content.

\textbf{Future directions.} Three avenues merit investigation: (1) targeted ablation at saturation layers to reduce susceptibility while preserving quality understanding; (2) extended anchor studies with odd, fractional, and non-numeric cues; (3) causal mediation analysis combining probing with intervention.

\appendix

\section{Supplementary Tables}\label{supplementary-tables}

\begin{table}[H]
\centering
\fittable{%
\begin{tabular}{@{}lccccc@{}}
\toprule
Model & Score (5b) & Fusion (5c) & Anchor bt. (5) & Anchor sat. (5) & $R^2$ best (5b) \\
\midrule
Gemma-3 & L1 (0.62) & L1 (0.98) & L12 (98.3\%) & L34 (99.95\%) & L34 (0.85) \\
Gemma-4 & L1 (0.72) & L2 (0.99) & L6 (99.0\%) & L12 (99.98\%) & L29 (0.91) \\
Qwen3.5 & L12 (0.57) & --- & L8 (99.8\%) & L12 (100\%) & L20 (0.69) \\
Qwen3-VL-4B & L4 (0.55) & --- & L7 (98.5\%) & L14 (100\%) & L21 (0.85) \\
MiniCPM-V-4 & L4 (0.65) & --- & L4 (93.8\%) & L12 (100\%) & L16 (0.81) \\
\bottomrule
\end{tabular}}
\caption{Cross-phase timing across five probed models. Score = score probing breakthrough, Fusion = fusion layer, Anchor bt.\ = anchor classification breakthrough, Anchor sat.\ = anchor saturation layer.}
\label{tab:cross-phase}
\end{table}


\begin{thebibliography}{99}

\bibitem{bleeker2024}
Bleeker~M.\,J.\,R., Hendriksen~M., Yates~A., de~Rijke~M., \emph{Demonstrating and Reducing Shortcuts in Vision-Language Representation Learning}, Transactions on Machine Learning Research, 2024.

\bibitem{cheng2025}
Cheng~H., Xiao~E., Wang~Y., Zhang~L., Zhang~Q., Cao~J., Xu~K., Sun~M., Hao~X., Gu~J., Xu~R., \emph{Exploring Typographic Visual Prompts Injection Threats in Cross-Modality Generation Models}, arXiv:2503.11519, 2025.

\bibitem{echterhoff2024}
Echterhoff~J., Liu~Y., Alessa~A., McAuley~J., He~Z., \emph{Cognitive Bias in Decision-Making with LLMs}, arXiv:2403.00811, 2024.

\bibitem{hufe2025}
Hufe~L., Venhoff~C., Purelku~E., Dreyer~M., Lapuschkin~S., Samek~W., \emph{Dyslexify: A Mechanistic Defense Against Typographic Attacks in CLIP}, arXiv:2508.20570, 2025.

\bibitem{li2025}
Li~Q., Ye~Z., Feng~X., Zhong~W., Ma~W., Feng~X., \emph{Causal Tracing of Object Representations in Large Vision Language Models: Mechanistic Interpretability and Hallucination Mitigation}, arXiv:2511.05923, 2025.

\bibitem{lou2024}
Lou~J., Sun~Y., \emph{Anchoring Bias in Large Language Models: An Experimental Study}, arXiv:2412.06593, 2024.

\bibitem{shi2025}
Shi~C., Yu~Y., Yang~S., \emph{Vision Function Layer in Multimodal LLMs}, arXiv:2509.24791, 2025.

\bibitem{steinberg2026}
Steinberg~J., Gal~O., \emph{Where Vision Becomes Text: Locating the OCR Routing Bottleneck in Vision-Language Models}, arXiv:2602.22918, 2026.

\bibitem{suri2023}
Suri~G., Slater~L.\,R., Ziaee~A., Nguyen~M., \emph{Do Large Language Models Show Decision Heuristics Similar to Humans? A Case Study Using GPT-3.5}, arXiv:2305.04400, 2023.

\bibitem{wang2024}
Wang~Z., Han~Z., Chen~S., Xue~F., Ding~Z., Xiao~X., Tresp~V., Torr~P., Gu~J., \emph{Stop Reasoning! When Multimodal LLM with Chain-of-Thought Reasoning Meets Adversarial Image}, arXiv:2402.14899, 2024.

\end{thebibliography}
\end{document}